\documentclass{article}

\usepackage[preprint]{neurips_2026}

\usepackage[utf8]{inputenc}
\usepackage[T1]{fontenc}
\usepackage{hyperref}
\usepackage{url}
\usepackage{booktabs}
\usepackage{amsfonts}
\usepackage{amsmath}
\usepackage{amssymb}
\usepackage{nicefrac}
\usepackage{microtype}
\usepackage{xcolor}
\usepackage{graphicx}
\usepackage{algorithm}
\usepackage{algpseudocode}
\usepackage{multirow}
\usepackage{subcaption}
\usepackage{enumitem}
\usepackage{wrapfig}
\usepackage{graphicx}
\usepackage{placeins}
\newcommand{\eg}{\textit{e.g.}}

\newcommand{\prism}{\textsc{PRISM}}

\definecolor{prismblue}{HTML}{2E6DA4}

\title{PRISM: Pareto-Efficient Retrieval over Intent-Aware Structured Memory for Long-Horizon Agents}

\author{%
  \begin{tabular}[t]{c}
    \bf Jingyi Peng\textsuperscript{1} \quad
    \bf Zhongwei Wan\textsuperscript{2} \quad
    \bf Weiting Liu\textsuperscript{3} \quad
    \bf Qiuzhuang Sun\textsuperscript{1*} \\[2pt]
    \normalfont\small \textsuperscript{1}Singapore Management University \enspace
    \textsuperscript{2}The Ohio State University \enspace
    \textsuperscript{3}Fudan University \\[1pt]
    \normalfont\small \textsuperscript{*}Corresponding author: \texttt{qzsun@smu.edu.sg}
  \end{tabular}%
}

\begin{document}

\maketitle

% =====================================================================
% ======================  章节输入区（已按顺序排列） ======================
% =====================================================================

 % =====================================================
%  ABSTRACT
% =====================================================

\begin{abstract}
    Long-horizon language agents accumulate conversation history far
    faster than any fixed context window can hold, making memory
    management critical to both answer accuracy and serving cost.
    Existing approaches either expand the context window without
    addressing what is retrieved, perform heavy ingestion-time fact
    extraction at substantial token cost, or rely on heuristic graph
    traversal that leaves both accuracy and efficiency on the table.
    We present \prism{}, a training-free retrieval-side framework that
    treats long-horizon memory as a joint retrieval-and-compression
    problem over a graph-structured memory. \prism{} combines four
    orthogonal inference-time components: \emph{Hierarchical Bundle
    Search} over typed relation paths, \emph{Query-Sensitive Edge
    Costing} that aligns traversal with detected query intent,
    \emph{Evidence Compression} that compresses the candidate bundle
    into a compact answer-side context, and \emph{Adaptive Intent
    Routing} that routes most queries through zero-LLM tiers. By
    formulating retrieval as min-cost selection over typed path
    templates and pairing it with an LLM-side compression step,
    \prism{} surfaces the right evidence under a strict context budget
    without any fine-tuning or modification to the upstream ingestion
    pipeline. Experiments on the LoCoMo benchmark show that \prism{}
    delivers substantially higher LLM-judge accuracy than every
    same-protocol baseline at an order-of-magnitude smaller context
    budget, occupying a previously empty corner of the
    accuracy--context-cost frontier and demonstrating a superior
    balance between answer quality and retrieval efficiency.
\end{abstract}
 
\section{Introduction}
\label{sec:intro}

In long-horizon agentic tasks involving multi-session conversations
and lifelong assistants \citep{maharana2024evaluating},
the effectiveness of large language model (LLM) agents is fundamentally
constrained by what information can be surfaced into the answer
model's context at query time. Because the attention window is finite
and degrades on long inputs \citep{liu2024lost}, an external
\emph{memory system} is required to store past experience and to
retrieve the relevant pieces when a new query arrives. The quality of
such a system is governed by two coupled quantities: the \emph{accuracy}
of the answers it supports, and the \emph{context cost}---the number of
tokens that must be packed into the answer model's prompt for every
query. High accuracy at low context cost is the regime that long-horizon
deployments care about, and pushing both quantities at once is the
central design challenge.

\begin{figure}[t]
      \centering
      \begin{subfigure}[b]{0.49\linewidth}
        \includegraphics[width=\linewidth]{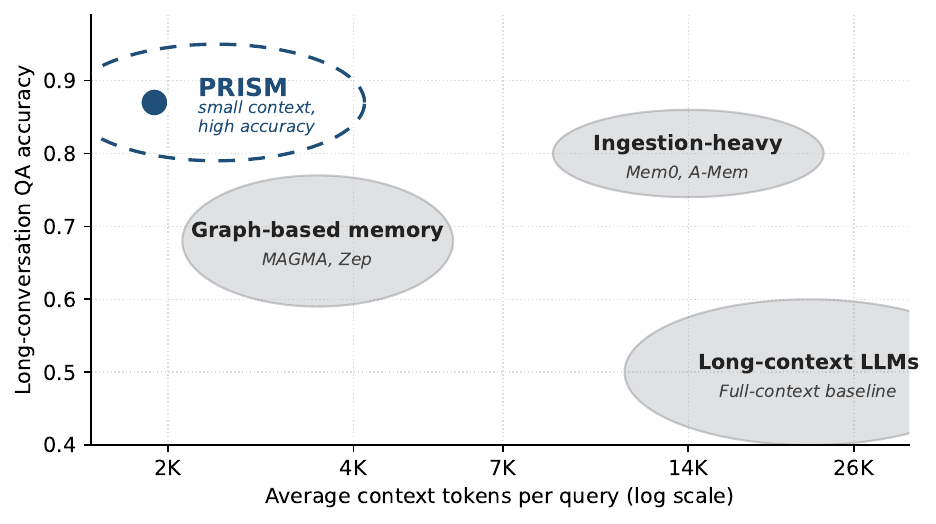}
        \caption{Accuracy--cost landscape}
        \label{fig:teaser-a}
      \end{subfigure}
      \hfill
      \begin{subfigure}[b]{0.49\linewidth}
        \includegraphics[width=\linewidth]{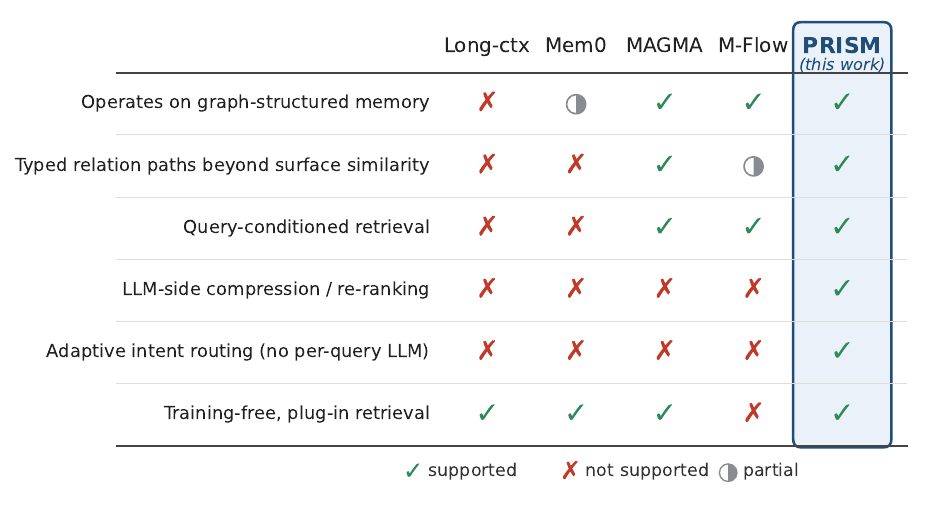}
        \caption{Design dimensions}
        \label{fig:teaser-b}
      \end{subfigure}
      \caption{(a)~Existing memory designs cluster in three regions of the
      accuracy--context-cost plane, leaving the high-accuracy /
      low-cost corner underfilled. (b)~PRISM is the only system that
      combines all six design dimensions we identify as relevant.}
      \label{fig:teaser}
      \vspace{-15pt}
\end{figure}

However, existing research has largely advanced these two quantities
along separate axes. Ingestion- and representation-centric systems
such as MemoryBank, A-Mem, Mem0, Zep, and MemoryOS
\citep{zhong2024memorybank, xu2025mem, chhikara2025mem0,
rasmussen2025zep, kang2025memory} focus on what to store, update,
and structure at write time, building richer memory units or
hierarchies that downstream retrieval can later draw from.
Retrieval- and context-management systems such as MemGPT, ReSum,
LightMem, and SimpleMem \citep{packer2023memgpt, wu2025resum,
fang2025lightmem, liu2026simplemem} instead reduce the active
context through memory tiers, periodic summarization, lightweight
filtering, or intent-aware retrieval planning. A third, more recent
line introduces explicit relational structure into memory: GraphRAG
and MAGMA \citep{edge2024local, jiang2026magma} build typed graphs
over events, entities, and causal links, and use graph traversal as
the retrieval primitive. A complementary direction trains the agent's
memory and retrieval behavior end-to-end with reinforcement learning
\citep{yu2026agentic}. Each line addresses a piece of the long-horizon
problem, but each leaves the other pieces unsolved.

As a consequence, the accuracy--context-cost plane is populated
unevenly. As illustrated in Figure~\ref{fig:teaser}(a), long-context
baselines pay a very high token cost for moderate accuracy;
ingestion-heavy systems can reach high accuracy but tend to retrieve
large candidate pools, again inflating context; graph-based memory
adds relational structure but, in current designs, sits at moderate
accuracy with moderate cost. The high-accuracy / low-cost corner is
conspicuously underfilled. We argue this gap is not a coincidence.
Reaching this corner requires \emph{simultaneously} retrieving along
the right relation type rather than by surface similarity, and
controlling how the retrieved candidates are condensed before they
reach the answer model. Existing systems address these two levers
separately---graph systems focus on the former but feed candidates
to the answer model essentially as-is, while compression-oriented
systems focus on the latter but operate over flat or weakly relational
memory. Recent improvements along both axes increasingly rely on
training a query-time policy or re-ingesting memory under a new
representation, which is expensive and tightly couples retrieval
logic to a specific backend. The high-accuracy / low-cost corner
therefore remains an open challenge.

To address this gap, we propose \textbf{PRISM}, a training-free
retrieval pipeline that occupies the empty corner by composing
typed-path retrieval with answer-side compression over an
already-constructed graph memory. PRISM's memory is built once at
write time via schema-guided extraction and consolidation, exposing
six typed edge types over a hierarchy of entities, facets, facet
points, and episodes. At query time, PRISM routes the query by intent
to a small set of relevant edge types, scores candidates along typed
relation paths rather than flat similarity, and applies an LLM-side
compression pass that re-ranks and condenses the candidate pool
before it is handed to the answer model. The retrieval pipeline
itself is training-free---no policy is learned and no fine-tuning
is required---and plugs into any backend that exposes the same edge
schema. Figure~\ref{fig:teaser}(b) summarizes the design dimensions
that PRISM combines and that prior systems leave individually
unaddressed.

We evaluate PRISM on LoCoMo \citep{maharana2024evaluating}, a
long-conversation QA benchmark covering single-hop, multi-hop,
temporal, and open-domain reasoning, and show that PRISM consistently
outperforms strong baselines while using an order of magnitude fewer
context tokens per query. Comprehensive ablations confirm that both
typed relation paths and answer-side LLM compression are necessary,
not substitutable: removing either lever collapses the system back
toward the regions that prior designs already occupy.

Our main contributions are as follows:
\begin{itemize}[leftmargin=1.5em, itemsep=2pt, topsep=2pt]
    \item We frame long-conversation memory as an
          \emph{accuracy--context-cost frontier} and identify the
          high-accuracy / low-cost corner that prior designs---across
          ingestion-heavy, retrieval-management, and graph-structured
          lines---leave largely empty.
    \item We propose \textbf{PRISM}, a training-free retrieval
          pipeline that combines intent-routed typed-path retrieval
          with answer-side LLM compression over a typed graph memory.
          To our knowledge, PRISM is the first system to combine
          these two levers without training a query-time policy.
    \item On LoCoMo, PRISM reaches $0.831$ LLM-judge score with
          roughly $2$K context tokens per query, beating every
          same-protocol baseline on overall accuracy while using
          $\sim$$13\times$ fewer tokens than a full-context answer
          model. Ablations isolate each module's contribution and
          identify answer-side LLM compression as the dominant
          token-reduction lever.
\end{itemize}

 % =====================================================================
% SECTION 2 — RELATED WORK
% =====================================================================
\section{Related Work}
\label{sec:related}

\paragraph{Long-context LLMs and KV-cache compression.}
One line scales the answer model itself to ingest longer inputs.
Frontier models now support windows of hundreds of thousands to
millions of tokens~\citep{team2024gemini,achiam2023gpt,anthropic2024claude}, and a growing
literature reduces inference cost through KV-cache eviction and
dynamic allocation~\citep{wan2024d2o,wan2024look,wan2025meda}.
However, these methods leave the retrieval question untouched: every
query still pays attention cost over the entire history, and large
models attend less reliably to evidence buried in long
inputs~\citep{liu2024lost}. \emph{In contrast}, PRISM operates on the
retrieval side and is orthogonal to long-context backbones, which can
consume its compact context more cheaply still.

\paragraph{Memory-augmented agents.}
A second line builds explicit memory artefacts over the conversation.
MemGPT~\citep{packer2023memgpt} pages information through a managed context
hierarchy under LLM control; A-MEM~\citep{xu2025mem} stores chunks as
Zettelkasten-style notes with semantic links; Zep~\citep{rasmussen2025zep}
maintains a bi-temporal knowledge graph; MAGMA~\citep{jiang2026magma} traverses
four orthogonal edge-graphs with a heuristic intent-aware beam search;
and Mem0 and Mem0$^{g}$~\citep{chhikara2025mem0} extract structured facts at
ingestion time and retrieve aggressively at query time. However, these
systems either keep memory as a flat node collection and recover
relations implicitly through similarity, or impose graph structure but
rely on heuristic traversal rules with per-query budgets in the
$3$--$7$K-token range. \emph{In contrast}, PRISM formalises retrieval
as a min-cost selection over typed path templates and follows it with
an LLM-side compression step that brings per-query context to
$\sim$$2$K tokens without sacrificing accuracy.

\paragraph{Retrieval-side optimisation.}
A third line treats memory as a retrieval problem and optimises the
\emph{selection} of evidence sent to the answer model. Dense
retrievers~\citep{karpukhin2020dense,khattab2020colbert} and lexical
baselines~\citep{robertson2009probabilistic} score passages by surface or embedding
similarity; cross-encoder rerankers~\citep{nogueira2019passage}
re-score a shortlist with a query-conditioned model; and
prompt-compression methods such as LLMLingua~\citep{jiang2023llmlingua} prune
tokens after retrieval. However, these methods are benchmarked on
flat document collections and cannot exploit indirect evidence linked
by causal, temporal, or evolution relations. \emph{In contrast},
PRISM brings retrieval-side discipline to graph-structured memory by
combining typed-path candidate generation, intent-conditioned edge
costs, and an LLM-based reranker over the candidate bundle.
 % =====================================================================
% SECTION 3 — PRISM DESIGN
% =====================================================================
\section{PRISM Design}
\label{sec:method}

% --------------------------------------------------------------------
\subsection{Architectural Overview}
\label{sec:overview}

\begin{figure}[t]
    \centering
    \includegraphics[width=\linewidth]{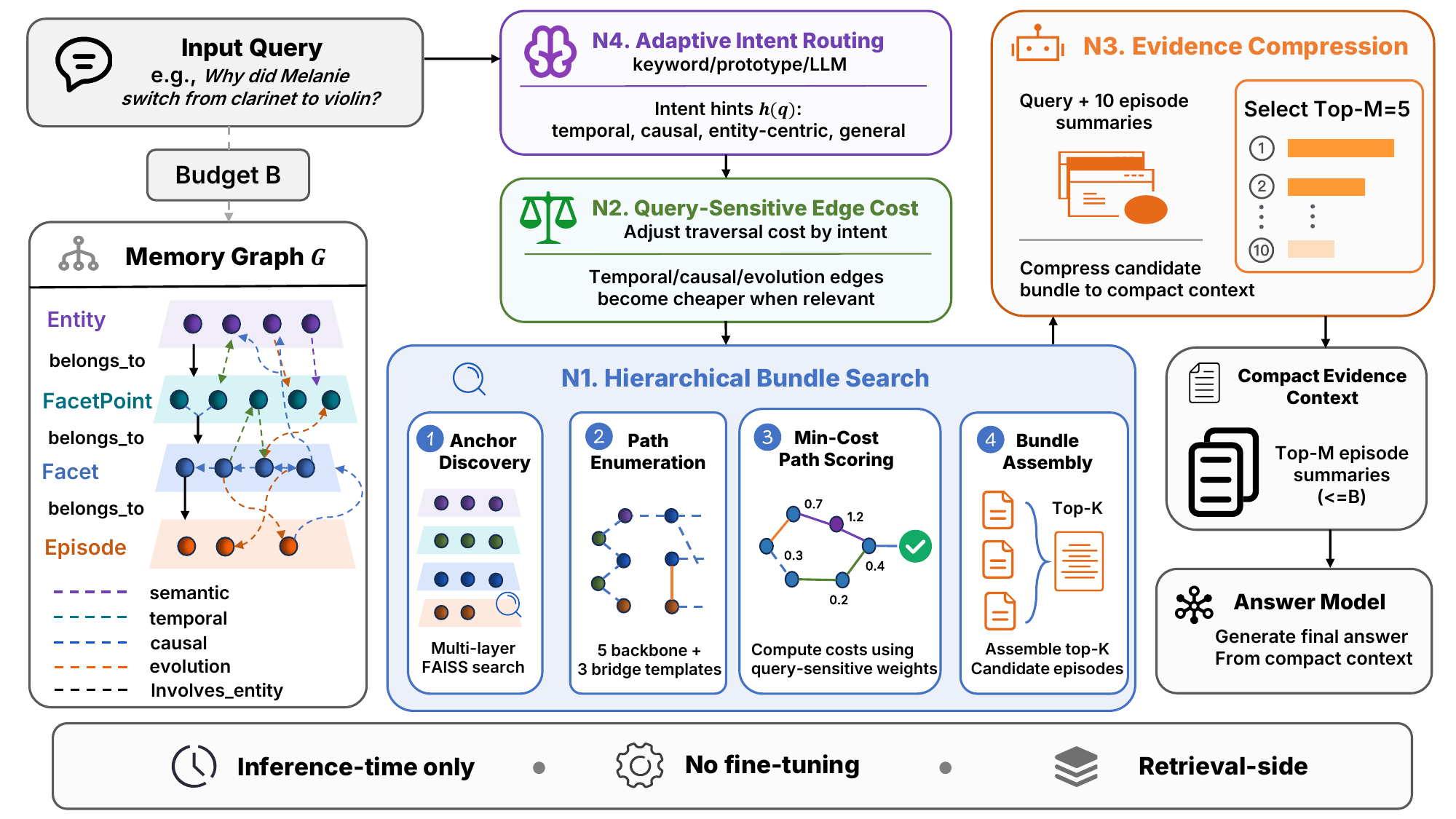}
    \caption{Architectural overview of PRISM.
    PRISM is composed of a four-layer memory graph and four inference-time modules:
    (1) N4 routes query intent; (2) N2 adjusts traversal costs over typed edges;
    (3) N1 searches relation paths and assembles candidate bundles; and
    (4) N3 compresses retrieved evidence into a compact context for the answer model.}
    \label{fig:overview}
    \vspace{-8pt}
\end{figure}

PRISM is a training-free, retrieval-side framework that operates over an
existing graph-structured memory $\mathcal{G}$ produced by any ingestion
pipeline. Given a query $q$ and a context budget $B$, PRISM aims to
return a set of conversation chunks (\emph{Episodes}) of total length at
most $B$ that support a correct downstream answer.
Figure~\ref{fig:overview} summarises the architecture: four
inference-time modules layer on top of the memory graph and compose
sequentially through data dependencies. N4 predicts query intent; N2
uses this intent to re-weight typed edges; N1 searches typed relation
paths to form a top-$K$ candidate bundle; and N3 compresses the bundle
to top-$M$ Episodes via a single LLM call. All modules are training-free
and apply unchanged to any backend exposing the node and edge schema
below. The remainder of this section defines the graph
(\S\ref{sec:graph}), develops each module
(\S\ref{sec:n1}--\S\ref{sec:n4}), and summarises the end-to-end
procedure (\S\ref{sec:pipeline}).

% --------------------------------------------------------------------
\subsection{Memory Graph and Notation}
\label{sec:graph}

PRISM retrieves over a multi-relational memory graph
$\mathcal{G} = (\mathcal{V}, \mathcal{E}, \tau)$ in which the node set
$\mathcal{V}$ is partitioned across four layers and the edge set
$\mathcal{E}$ across two structural families. The four node layers
form a hierarchy of granularity: an \textbf{Entity} is a named anchor
(person, place, object) shared across multiple chunks; a
\textbf{FacetPoint} is an atomic fact extracted from a single chunk;
a \textbf{Facet} groups thematically related FacetPoints within a
chunk; and an \textbf{Episode} is the original conversation chunk,
which is also the unit eventually returned to the answer model. The
edge set decomposes into \emph{hierarchical edges}
(\texttt{belongs\_to}, encoding containment
Entity\,$\to$\,FacetPoint\,$\to$\,Facet\,$\to$\,Episode) and
\emph{relation edges}, comprising five typed cross-cutting relations:
\emph{semantic}, \emph{temporal}, \emph{causal}, \emph{evolution},
and \emph{\texttt{involves\_entity}}. Each node $v$ carries a content
embedding $\mathbf{e}_v \in \mathbb{R}^d$ obtained at ingestion time,
each non-\texttt{belongs\_to} edge carries an analogous edge
embedding, and we write $\tau(e)$ for the type of an edge $e$ throughout. We
denote the query embedding by $\mathbf{q}$. All text embeddings are
$L_2$-normalised before indexing, so inner-product search with
\texttt{IndexFlatIP} is equivalent to cosine-similarity retrieval,
which we use as the distance throughout.

% --------------------------------------------------------------------
\subsection{N1: Hierarchical Bundle Search}
\label{sec:n1}

Flat vector retrieval treats memory as a bag of passages and selects
the top-$k$ closest to $\mathbf{q}$, which degrades on queries whose
evidence is spread across multiple chunks linked by typed relations
rather than surface similarity. The \textbf{Hierarchical Bundle
Search} module (N1) recovers such evidence by enumerating typed
\emph{path templates} from query anchors to candidate Episodes and
scoring each Episode by the minimum-cost path that reaches it.

\textbf{Anchor discovery.}
We retrieve a small pool of anchor nodes by issuing a layer-specific
top-$k$ vector search against each of the four node layers and taking
the union:
\begin{equation}
    \mathcal{A}
    = \bigcup_{\ell \in \{\text{Ep},\text{Fc},\text{FP},\text{En}\}}
      \mathrm{TopK}\bigl( F_\ell,\, \mathbf{q} \bigr),
    \label{eq:anchors}
\end{equation}
where $F_\ell$ denotes the FAISS index for layer $\ell$. Each anchor
$a \in \mathcal{A}$ is annotated with its \emph{anchor cost}
$d(a) = 1 - \cos(\mathbf{q}, \mathbf{e}_a)$.

\textbf{Path templates.}
A path template is a typed sequence of node and edge types ending at
an Episode. N1 enumerates eight templates: a \emph{backbone} family
over \texttt{belongs\_to} edges and a \emph{relation-bridge} family
that crosses one typed relation edge before re-entering the hierarchy.
The latter is the structural basis for answering multi-hop and causal
questions. Full enumeration is given in Appendix~\ref{app:path-templates}.

\textbf{Path costing and bundle assembly.}
For an instantiated path
$\pi = (a, e_1, v_1, \ldots, e_h, \mathrm{Ep})$ with $h$ hops, the
path cost is
\begin{equation}
    \mathrm{Cost}(\pi)
    = d(a)
    + \sum_{i=1}^{h} \bigl( c_{\text{edge}}(e_i)
                          + c_{\text{hop}} \bigr),
    \label{eq:path-cost}
\end{equation}
where $c_{\text{edge}}(e_i)$ is the per-edge traversal cost from
Query-Sensitive Edge Cost (\S\ref{sec:n2}) and $c_{\text{hop}}$ is a
small penalty preventing arbitrarily long paths. Each Episode is
scored by the \emph{minimum} cost over all paths terminating at it,
\begin{equation}
    s(\mathrm{Ep})
    \;=\;
    \min_{\pi \,\in\, \Pi(\mathrm{Ep})} \mathrm{Cost}(\pi),
    \label{eq:min-cost}
\end{equation}
encoding the inductive bias that a single strong evidence chain
suffices to establish relevance. The top-$K$ Episodes by ascending
$s(\mathrm{Ep})$ form the candidate \emph{bundle} passed to Evidence
Compression (\S\ref{sec:n3}), with the minimum-cost path retained as a
retrieval trace. For example, the query \emph{``Why did Melanie switch
from clarinet to violin?''} has low surface similarity to the chunk
\emph{``Melanie attended a violin recital that moved her''}, but a
causal-bridge path lifts this chunk into the bundle.

% --------------------------------------------------------------------
\subsection{N2: Query-Sensitive Edge Cost}
\label{sec:n2}

Edge types in a multi-relational memory graph are not interchangeable:
a \emph{when} query benefits most from temporal and evolution edges,
a \emph{why} query from causal edges. Treating all edges with a
uniform cost in Eq.~\ref{eq:path-cost} discards this signal. The
\textbf{Query-Sensitive Edge Cost} module (N2) re-weights the per-edge
cost conditional on a detected query intent
$h(q) \subseteq \{\textsc{temporal},\, \textsc{causal},\,
\textsc{multi\_hop},\, \textsc{entity\_centric},\, \textsc{general}\}$
supplied by Adaptive Intent Routing (\S\ref{sec:n4}); among these,
only \textsc{temporal} and \textsc{causal} trigger edge-cost
discounts, while \textsc{multi\_hop} serves a routing-side role
described in \S\ref{sec:n4}:
\begin{equation}
    c_{\text{edge}}(e_i)
    \;=\;
    \alpha\bigl(\tau(e_i),\, h(q)\bigr) \cdot
    \begin{cases}
        1 - \cos(\mathbf{e}_{e_i}, \mathbf{q})
            & e_i \in \mathcal{R}_{\text{FAISS}}, \\[2pt]
        c_0\bigl(\tau(e_i)\bigr)
            & \text{otherwise},
    \end{cases}
    \qquad \alpha \in (0, 1],
    \label{eq:edge-cost}
\end{equation}
where $\mathcal{R}_{\text{FAISS}}$ is the set of edges incident to
any anchor recalled by the FAISS top-$k$ search at any of the four
node layers, $\mathbf{e}_{e_i}$ is the edge embedding, $c_0(\cdot)$
is a per-type fallback cost applied only when $e_i$ falls outside
the FAISS recall, and the intent-conditioned discount factor
$\alpha$ encodes intent--edge match rules:
\begin{equation}
    \alpha\bigl(\tau, h(q)\bigr)
    \;=\;
    \begin{cases}
        0.5 & \tau = \textsc{temporal},\ \textsc{temporal} \in h(q),\\[1pt]
        0.5 & \tau = \textsc{causal},\ \textsc{causal} \in h(q),\\[1pt]
        0.7 & \tau = \textsc{evolution},\ \textsc{temporal} \in h(q),\\[1pt]
        1.0 & \text{otherwise}.
    \end{cases}
    \label{eq:alpha}
\end{equation}
The discount is multiplicative, so the adjustment is monotone in the
original cost ordering: a temporal-bridge path becomes roughly half
the cost of an otherwise identical backbone path \emph{when, and only
when,} the query carries a temporal intent. For example, the queries
\emph{``When did Melanie start violin?''} and \emph{``Why did Melanie
switch to violin?''} traverse the same memory subgraph but, under
Eq.~\ref{eq:alpha}, select different cheapest paths---one along the
temporal chain to a date-anchored chunk, the other along the causal
bridge to the recital chunk.

% --------------------------------------------------------------------
\subsection{N3: Evidence Compression}
\label{sec:n3}

Path cost is a structural signal: it identifies which Episodes lie on
cheap evidence chains, but is agnostic to their semantic content. An
Episode whose minimum-cost path passes through a generic anchor
(\eg, a popular entity) can therefore rank highly while being
topically irrelevant, and the leading context positions may be
determined by structural rather than semantic relevance. The
\textbf{Evidence Compression} module (N3) addresses this gap by issuing
a single LLM call that selects top-$M$ Episodes from the top-$K$ bundle
$\mathcal{B}$ on content alone:
\begin{equation}
    \mathcal{B}^{*}
    \;=\;
    \mathrm{LLMSelect}\bigl(q,\,
        \{\mathrm{Summary}(\mathrm{Ep}) : \mathrm{Ep} \in \mathcal{B}\},\,
        M \bigr),
    \quad
    |\mathcal{B}^{*}| = M,
    \label{eq:compress}
\end{equation}
where $\mathrm{Summary}(\mathrm{Ep})$ is the Episode summary prepared
at ingestion time. The selection prompt is content-only: it sees
Episode summaries but neither path-cost scores nor bundle metadata,
so its judgement is independent of the structural signal that produced
the candidate set. N3 thus acts both as a precision filter that drops
structurally cheap but topically irrelevant Episodes, and as a context
compressor that shrinks the answer-side context from $K$ to $M$
Episodes ($M < K$). For example, under \emph{``What instruments does
Melanie play?''}, a chunk about her cousin's piano recital may share
popular anchors with the query and earn a low path cost, but
$\mathrm{LLMSelect}$ drops it as off-topic.

% --------------------------------------------------------------------
\subsection{N4: Adaptive Intent Routing}
\label{sec:n4}

Hierarchical Bundle Search and Query-Sensitive Edge Cost both depend on
the intent label $h(q)$. The natural implementation issues one LLM
classification call per query, but for a system whose central claim is
\emph{context efficiency}, a per-query classifier-side LLM call is
itself a measurable cost. The \textbf{Adaptive Intent Routing} module
(N4) routes each query through a three-tier cascade ordered from
cheapest to most expensive:
\begin{equation}
    h(q)
    \;=\;
    \begin{cases}
        h_{\text{kw}}(q)
            & \mathrm{KeywordMatch}(q),\\[1pt]
        h_{\text{proto}}(q)
            & \text{else if } \max_p \cos(\mathbf{q}, \mathbf{p})
                     \,>\, \theta_{\text{proto}},\\[1pt]
        h_{\text{LLM}}(q)
            & \text{otherwise}.
    \end{cases}
    \label{eq:cascade}
\end{equation}
The first tier matches high-precision regular expressions on intent
triggers (\eg, \emph{when}/\emph{before} for \textsc{temporal},
\emph{why}/\emph{because} for \textsc{causal}); the second compares
$\mathbf{q}$ against a bank of intent prototype embeddings precomputed
offline; the third invokes an LLM only when both cheap tiers fail.
Queries whose cascade produces no confident label are assigned
$h(q) = \emptyset$, leaving Eq.~\ref{eq:edge-cost} unchanged. N4
emits five labels: \textsc{temporal} and \textsc{causal} trigger
edge-cost discounts in Eq.~\ref{eq:alpha}; \textsc{multi\_hop}
suppresses spurious \textsc{entity\_centric} labels on inference-heavy
queries but does not itself discount edges; \textsc{entity\_centric}
is a routing label with no edge-cost effect, and \textsc{general} is
the safe-default label when no confident intent is recovered. For
instance, \emph{``When did Melanie start violin?''} is dispatched at
the keyword tier, while \emph{``What changed for Melanie around the
spring?''} reaches the LLM fallback---concentrating LLM cost on hard
cases.

% --------------------------------------------------------------------
\subsection{End-to-End Inference}
\label{sec:pipeline}

The four modules compose into a fixed inference sequence:
$h(q) \leftarrow$ Adaptive Intent Routing,
$c_{\text{edge}} \leftarrow$ Query-Sensitive Edge Cost,
$\mathcal{B} \leftarrow$ Hierarchical Bundle Search,
$\mathcal{B}^{*} \leftarrow$ Evidence Compression. The pipeline
issues at most two LLM calls per query in the worst case (the
Adaptive Intent Routing fallback and the Evidence Compression
selection), and only one call on queries handled by the cheap tiers
of Adaptive Intent Routing; all other operations are deterministic
vector arithmetic and graph traversal. Full pseudocode is given in
Algorithm~\ref{alg:prism} (Appendix~\ref{app:algorithm}).
 \section{Experiments}
\label{sec:expert}

% =====================================================================
\subsection{Experimental Setup}
\label{sec:setup}

\paragraph{Benchmark and protocol.}
We evaluate on LoCoMo~\citep{maharana2024evaluating}, using the standard
10-conversation split and categories 1--4 (single-hop, multi-hop, temporal,
and open-domain), for a total of $1{,}540$ QA pairs; category~5 is excluded
because it tests adversarial refusal rather than retrieval. Unless otherwise
stated, all same-protocol rows use \texttt{gpt-4o-mini} as both answer and
judge model with \texttt{temperature}$=0.0$, and share the same answer prompt,
judge prompt, tokenizer, and token-counting procedure.

\paragraph{Baselines and references.}
We compare PRISM against Full Context, MAGMA~\citep{jiang2026magma},
Mem0 and Mem0$^{g}$~\citep{chhikara2025mem0}, M-Flow~\citep{mflow2026},
and the Mem0 commercial platform~\citep{chhikara2025mem0}. Mem0$^{g}$ is
the closest direct competitor, as it also exploits an explicit conversation
graph. Full Context, MAGMA, Mem0, Mem0$^{g}$, PRISM, and its ablations are
all re-run under our exact protocol. M-Flow, PRISM (gpt-5.5), and the
Mem0 platform are reported as different-protocol references: M-Flow and
the Mem0 platform are taken from their published configurations, while
for PRISM (gpt-5.5) we hold the retrieval pipeline and judge prompt fixed
and only swap \texttt{gpt-5.5} as the answer model on the same ingest
checkpoint as PRISM.

\paragraph{Metrics.}
We report three metrics in Table~\ref{tab:locomo_main}: (i) the
\emph{LLM-judge score}, defined as
$\text{CORRECT}/(\text{CORRECT}{+}\text{WRONG})$ over the $1{,}540$ questions;
(ii) \emph{context tokens per query}, a retrieval-cost proxy measured with
a fixed tokenizer and held identical across all same-protocol rows so that
budgets are directly comparable; and (iii) \emph{per-1K efficiency}, defined
as the judge score per $1$K retrieved context tokens. Component analysis in
\S\ref{sec:ablation} additionally uses Evidence Recall@$K$ (ER@$K$), and
paired ablations report McNemar mid-$p$ values and $95\%$ paired-bootstrap
confidence intervals ($2{,}000$ resamples). Full prompts and statistical
details are in Appendix~\ref{app:prompt-judge} and
Appendix~\ref{app:ci_full}.
% =====================================================================
\subsection{Main Results on LoCoMo}
\label{sec:main}
% =====================================================================

\begin{table*}[t]
\centering
\caption{Results on LoCoMo (categories 1--4, $1{,}540$ QA pairs). Best same-protocol result per column in \textbf{bold}; the different-protocol group is reported for reference only.}
\label{tab:locomo_main}
\footnotesize
\setlength{\tabcolsep}{3.5pt}
\renewcommand{\arraystretch}{0.95}

\resizebox{\textwidth}{!}{
\begin{tabular}{@{}lccccccc@{}}
\toprule
Method & Multi-Hop & Temporal & Open-Domain & Single-Hop & Overall & Per-1K Eff. & Ctx tokens/query \\
\midrule
\multicolumn{8}{@{}l}{\emph{Same-Protocol (gpt-4o-mini Answer Model)}} \\
Full Context     & 0.468 & 0.562 & 0.486 & 0.630 & 0.481 & 0.018 & 26{,}031 \\
MAGMA            & 0.528 & 0.650 & 0.517 & 0.776 & 0.688 & 0.204 & 3{,}370 \\
Mem0             & 0.512 & 0.555 & 0.729 & 0.671 & 0.669 & 0.379 & \textbf{1{,}764} \\
Mem0$^{g}$       & 0.472 & 0.581 & 0.757 & 0.657 & 0.684 & 0.189 & 3{,}616 \\
\textbf{PRISM (ours)}
                 & \textbf{0.787} & \textbf{0.788} & \textbf{0.813} & \textbf{0.863} & \textbf{0.831} & \textbf{0.411} & 2{,}023 \\
\midrule
\multicolumn{8}{@{}l}{\emph{Different-Protocol (Stronger Answer LLM / Managed Commercial Pipeline)}} \\
M-Flow           & 0.752 & 0.794 & 0.583 & 0.876 & 0.818 & 0.316 & 2{,}588 \\
\textbf{PRISM (gpt-5.5)}
                 & 0.890 & 0.879 & 0.927 & 0.892 & 0.891 & 0.440 & 2{,}023 \\
Mem0 platform    & --    & --    & --    & --    & 0.916 & 0.131 & $\sim$7{,}000 \\
\bottomrule
\end{tabular}
}

\vspace{-6pt}
\end{table*}

Table~\ref{tab:locomo_main} reports LoCoMo cat1--4 results,
partitioned by evaluation protocol.

\paragraph{PRISM Wins All Same-Protocol Comparisons.}
Within the same-protocol group, PRISM achieves the highest overall
LLM-Judge score ($0.831$), outperforming Mem0$^{g}$ by
$+14.2$\,pp, MAGMA by $+14.0$\,pp, and Full Context by
$+35.2$\,pp. PRISM also wins every per-category column, making it
the sole same-protocol method on the accuracy--efficiency frontier.
Notably, the smallest-context baseline (Mem0 at $1{,}764$ tokens)
trails PRISM by $16.2$\,pp despite a tighter budget: aggressive
context compression alone is insufficient---retrieval quality is the
binding constraint. PRISM also delivers the best accuracy per token,
attaining $0.411$ judge points per $1$K retrieved tokens versus
$0.379$ for the next-best Mem0.

\paragraph{Retrieval Quality Beats Raw Context Size.}
Feeding all $\sim$$26$K tokens of a LoCoMo dialogue directly to
\texttt{gpt-4o-mini} attains only $0.481$ judge---below \emph{every}
retrieval-based method, including the simplest one (Mem0 at
$0.669$). For long-conversation QA on this benchmark, \emph{what} is
retrieved matters substantially more than \emph{how much}, even when
the underlying answer model can in principle ingest the whole
dialogue.

\paragraph{Different-Protocol References.}
The different-protocol group includes three reference points.
M-Flow attains $0.818$ overall---$1.2$\,pp below PRISM at
$128\%$ of PRISM's context budget---despite using the stronger
\texttt{gpt-5-mini} answer model; per category, M-Flow leads on
Temporal and Single-Hop while PRISM leads on Multi-Hop ($+3.5$\,pp)
and Open-Domain ($+22.9$\,pp). Swapping the answer model in PRISM
itself from \texttt{gpt-4o-mini} to \texttt{gpt-5.5}---holding the
retrieval pipeline and ingest checkpoint fixed---lifts the overall
judge score from $0.831$ to $0.891$ ($+6.0$\,pp) at the same
$2{,}023$-token context budget, indicating that the residual
errors of PRISM are partly answer-model-bound rather than
retrieval-bound. Finally, the Mem0 commercial platform reaches
$0.916$ overall but at $\sim$$3.4\times$ PRISM's context budget and
under a managed pipeline we cannot replicate; its per-1K efficiency
($0.131$) is roughly one-third that of PRISM. Together, these three
references show that absolute accuracy on LoCoMo can be improved by
a stronger answer model or a larger budget, while PRISM occupies a
previously empty corner of the accuracy--efficiency frontier
within the same-protocol regime.

% =====================================================================
\subsection{Context-Efficiency Frontier}
\label{sec:pareto}
% =====================================================================

\begin{wrapfigure}[14]{r}{0.52\textwidth}
    \vspace{-10pt}
    \centering
    \includegraphics[width=\linewidth]{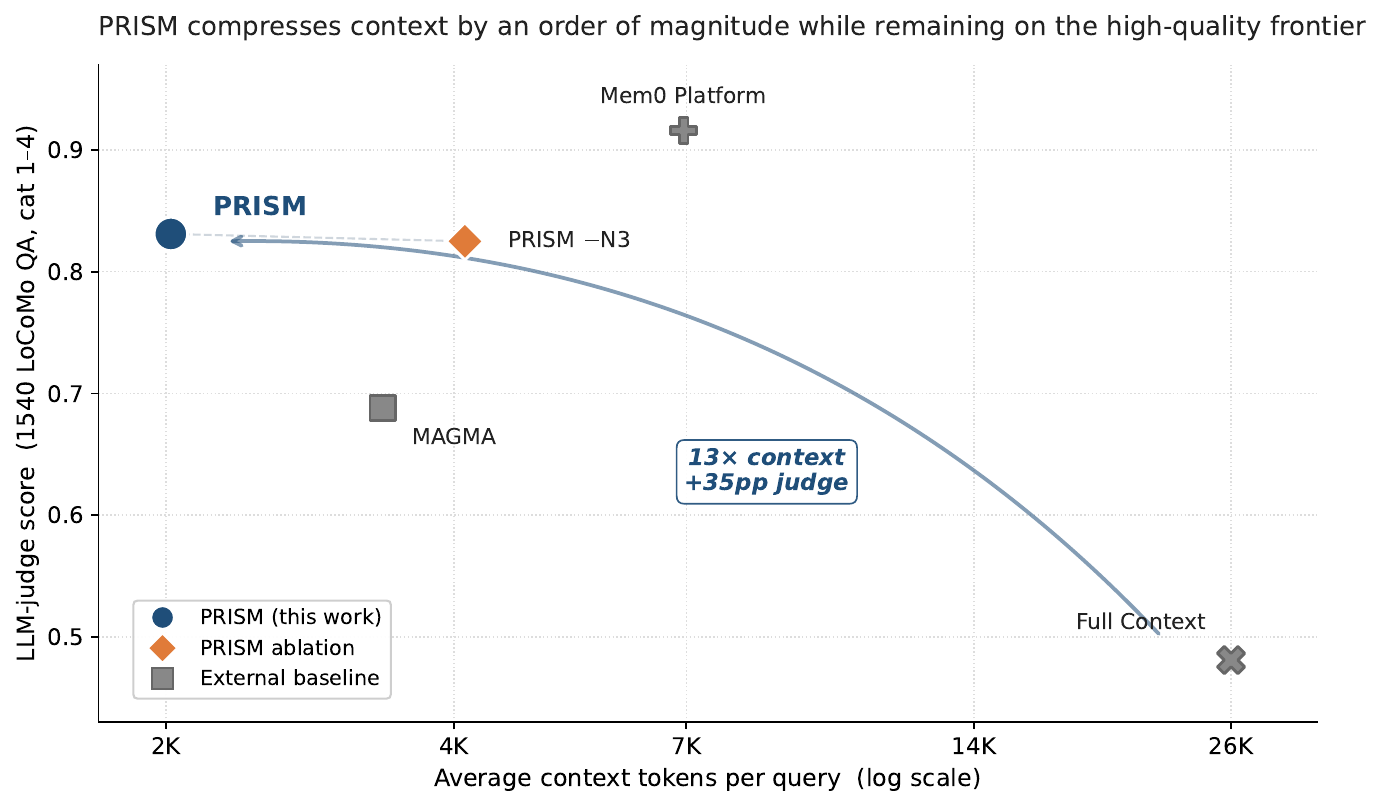}
    \caption{Accuracy--context trade-off on LoCoMo. Each point is one
    system; $x$-axis is average retrieved context tokens per query,
    $y$-axis is LLM-judge score.}
    \label{fig:pareto}
    \vspace{-18pt}
\end{wrapfigure}

Figure~\ref{fig:pareto} places PRISM on the accuracy--context plane
against representative baselines and ablations spanning the LoCoMo
memory-strategy spectrum, from no-retrieval to a managed commercial
pipeline. PRISM sits in the upper-left corner: high judge at low
context.

\paragraph{An Order-of-Magnitude Reduction in Context.}
Compared with Full Context, PRISM achieves a $\mathbf{13\times}$
context reduction ($\sim$$26$K $\to$ $\sim$$2$K tokens) with a
$\mathbf{+35}$\,pp judge gain ($0.481 \to 0.831$). The two
improvements are aligned rather than traded off: structured candidate
generation surfaces compact evidence candidates, and content-only
re-ranking (N3) filters the bundle by topical relevance.

\paragraph{Evidence Compression Sets the Corner.}
The orange diamond (\textsc{PRISM}\,$-$\,N3) isolates Evidence
Compression's contribution. Without N3, PRISM passes the top-$10$
candidate bundle directly to the answer model, roughly doubling the
per-query context while moving judge by less than one point. N3 is
therefore the structural reason PRISM sits at the corner of the
frontier rather than along its slope; the detailed ablation is reported
in \S\ref{sec:ablation}.

\paragraph{No Same-Protocol Method Matches PRISM's Budget.}
MAGMA is fully dominated on this plane, with lower judge at
$1.7\times$ the context. The Mem0 commercial platform reaches higher
accuracy, but under a different protocol, at $\sim$$3.4\times$
PRISM's context and roughly one-third its per-1K efficiency. Within
the same-protocol regime, no prior method delivers comparable
accuracy at PRISM's context budget.

\FloatBarrier

% =====================================================================
\subsection{Ablation Study}
\label{sec:ablation}
% =====================================================================

% -------------------------------------------------------------------------
% Table 3 -- PRISM component ablation on LoCoMo.
% -------------------------------------------------------------------------
\begin{table*}[t]
    \centering
    \small
    \setlength{\tabcolsep}{4.5pt}
    \caption{Component ablation on LoCoMo. Each row changes one flag relative to PRISM; bold marks large deviations from PRISM.}
    \label{tab:ablation}
    \begin{tabular}{l|c|ccccc|c}
        \toprule
        & \textbf{Judge} & \multicolumn{5}{c|}{\textbf{ER@5}} & \textbf{Ctx.} \\
        \cmidrule(lr){3-7}
        Configuration & Overall & Overall & Multi-hop & Temporal & Open-dom. & Single-hop & Tokens \\
        \midrule
        \textbf{PRISM}
            & $0.831$ & $0.694$ & $0.456$ & $0.781$ & $0.464$ & $0.766$ & $2{,}023$ \\
        \midrule
        \quad $-$\,N1 (relation paths)
            & $0.831$ & $0.694$ & $0.457$ & $0.780$ & $0.468$ & $0.766$ & $2{,}024$ \\
        \quad $-$\,N2 (cost adjustment)
            & $0.831$ & $0.694$ & $0.456$ & $0.783$ & $0.459$ & $0.766$ & $2{,}020$ \\
        \quad $-$\,N3 (LLM rerank)
            & $0.825$ & $\mathbf{0.627}^{\star}$ & $\mathbf{0.363}^{\star}$ & $\mathbf{0.737}^{\star}$ & $\mathbf{0.395}^{\star}$ & $\mathbf{0.699}^{\star}$ & $\mathbf{4{,}108}$ \\
        \quad $+$\,N4 (hybrid intent)
            & $0.833$ & $0.694$ & $0.456$ & $0.778$ & $0.464$ & $0.768$ & $2{,}023$ \\
        \bottomrule
    \end{tabular}

\end{table*}

Table~\ref{tab:ablation} evaluates four single-flag variants of
PRISM: disabling relation paths (N1), query-sensitive edge costs
(N2), or LLM re-ranking (N3); and enabling the optional adaptive
intent routing (N4). All variants use the same ingest checkpoint,
retrieval pipeline, answer prompt, answer model, and judge prompt;
only the indicated flag changes. We report LLM-judge accuracy,
Evidence Recall@$5$ (ER@$5$), and average answer-side context tokens.
Full paired-bootstrap CIs and McNemar mid-$p$ values are reported in
Appendix~\ref{app:ci_full}.

\paragraph{N3 Is the Main Driver of Both Precision and Compression.}
Removing the LLM re-ranker is the only ablation that materially
changes retrieval quality. ER@$5$ falls from $0.694$ to $0.627$
($-6.8$\,pp), with consistent drops across all four categories
(multi-hop $-9.3$, temporal $-4.4$, open-domain $-6.9$, single-hop
$-6.8$). The effect is even larger at top-$1$, where ER@$1$ improves
from $0.342$ without re-ranking to $0.551$ with re-ranking
($+20.8$\,pp). Judge accuracy moves only from $0.831$ to $0.825$
(paired-bootstrap CI overlapping zero), so Evidence Compression's
primary effect is on \emph{which} evidence reaches the top of the
context rather than on final answer accuracy. Disabling N3 also
inflates the answer-side context from $2{,}023$ to $4{,}108$ tokens,
confirming Evidence Compression as the dominant context-compression
component.

\paragraph{N1 and N2 Are Null on LoCoMo, but Targeted at Bridge-Style
Settings.}
Disabling relation paths or query-sensitive edge costs leaves ER@$5$,
judge accuracy, and context length essentially unchanged. We verified
that this is not an artefact of the LLM re-ranker masking earlier
differences by recomputing ER@$K$ at the pre-rerank candidate-pool
level for $K \in \{1, 3, 5, 10, 20\}$: PRISM and the corresponding
ablations produced bit-identical candidate sets for $1{,}536$ of
$1{,}540$ questions (maximum ER@$K$ difference $0.0007$). To diagnose
this null result, we manually annotated a random sample of $50$
multi-hop questions (full table in Appendix~\ref{app:multihop_anno}):
$72\%$ are fully anchor-discoverable (each gold turn shares a clear
topical or named-entity anchor with the question), and only
$\mathbf{6\%}$ (Wilson $95\%$ CI $[2.1\%, 16.5\%]$) require the kind
of explicit two-hop entity bridge that N1 and N2 are designed to
support. Combined with the fact that $73.4\%$ of category 1--4
questions cite a single evidence entry, the bridging cases that
N1/N2 actively target make up $\lesssim 3\%$ of the benchmark---too
small to produce a measurable judge-score delta. The takeaway is
that LoCoMo is mostly anchor-discoverable; we expect N1 and N2 to
matter more on benchmarks with lexically mismatched multi-hop
evidence such as MuSiQue~\citep{trivedi2022musique} or bridge-style
HotpotQA~\citep{yang2018hotpotqa}.

\paragraph{N4 Reduces Classifier-Side LLM Calls at No Accuracy
Cost.}
Adaptive Intent Routing is an optional efficiency extension whose
contribution is on the cost axis rather than the accuracy axis.
Table~\ref{tab:routing} reports the routing breakdown: across the
$1{,}540$ questions, $\mathbf{42.3\%}$ are dispatched through no-LLM
paths (\texttt{keyword\_gated}, \texttt{prototype}, or \texttt{none}),
with judge scores comparable to the LLM path ($0.829$ vs.\ $0.838$).
Savings concentrate on temporal queries, where $82.6\%$ match keyword
triggers, while only $8$--$23\%$ of multi-hop, single-hop, and
open-domain queries do; the per-category breakdown is reported in
Appendix~\ref{app:routing_per_category}. Whether this $42.3\%$
saving is statistically free is the subject of
Table~\ref{tab:paired}, which compares PRISM against PRISM+N4 on the
same questions: the two configurations disagree on only $88$ of
$1{,}540$ items, yielding $\Delta = +0.26$\,pp with McNemar mid-$p$
$0.71$ and a $95\%$ paired bootstrap CI of $[-0.91, +1.43]$\,pp; all
per-category CIs contain zero and all $p$-values exceed $0.45$. We
therefore do not claim N4 improves accuracy---its role is to remove
classifier-side LLM calls while preserving the same accuracy
profile.

% -------------------------------------------------------------------------
% Tables 3--4 -- N4 routing and paired comparison.
% -------------------------------------------------------------------------
\begin{table*}[t]
    \centering
    \footnotesize

    \begin{minipage}[t]{0.36\textwidth}
        \vspace{0pt}
        \centering
        \caption{N4 routing breakdown on LoCoMo. }
        \label{tab:routing}
        \setlength{\tabcolsep}{3.8pt}
        \begin{tabular}{@{}lrrcc@{}}
            \toprule
            Path & Share & $n$ & Judge & LLM \\
            \midrule
            \texttt{kw-gated}  & $32.2\%$ & $496$ & $0.829$ & No \\
            \texttt{prototype} & $\phantom{0}8.6\%$ & $132$ & $0.811$ & No \\
            \texttt{llm}       & $57.7\%$ & $889$ & $0.838$ & Yes \\
            \texttt{none}      & $\phantom{0}1.5\%$ & $\phantom{0}23$ & $0.870$ & No \\
            \midrule
            \textbf{No-LLM} & \textbf{42.3\%} & \textbf{651} & $0.829$ & --- \\
            \bottomrule
        \end{tabular}
    \end{minipage}
    \hfill
    \begin{minipage}[t]{0.61\textwidth}
        \vspace{0pt}
        \centering
        \caption{Paired accuracy comparison between PRISM and PRISM+N4. $\Delta$ is in pp.}
        \label{tab:paired}
        \setlength{\tabcolsep}{3.2pt}
        \begin{tabular}{@{}lrrrrrl@{}}
            \toprule
            Category & $n$ & PRISM & +N4 & $\Delta$ & McN. & 95\% CI \\
            \midrule
            Overall      & $1540$ & $0.8305$ & $0.8331$ & $+0.26$ & $0.71$ & $[-0.91,+1.43]$ \\
            \midrule
            Multi-hop    & $\phantom{0}282$ & $0.7872$ & $0.7801$ & $-0.71$ & $0.75$ & $[-3.90,+2.48]$ \\
            Temporal     & $\phantom{0}321$ & $0.7882$ & $0.7882$ & $\phantom{+}0.00$ & $0.92$ & $[-3.12,+3.12]$ \\
            Open-domain  & $\phantom{00}96$ & $0.8125$ & $0.8229$ & $+1.04$ & $0.86$ & $[-4.17,+6.25]$ \\
            Single-hop   & $\phantom{0}841$ & $0.8633$ & $0.8692$ & $+0.59$ & $0.45$ & $[-0.83,+1.90]$ \\
            \bottomrule
        \end{tabular}
    \end{minipage}

    \vspace{-6pt}
\end{table*}

\paragraph{Top-$5$ Is the Operating Elbow.}
Reducing the rerank pool from top-$5$ to top-$3$ (PRISM-top$3$)
lowers ER@$5$ by $2.3$\,pp without a measurable change in judge
accuracy, supporting top-$5$ as the operating elbow for Evidence
Compression. The complementary answer-model swap to
\texttt{gpt-5.5} is reported in Table~\ref{tab:locomo_main} and
discussed in \S\ref{sec:main} as a different-protocol reference.

\paragraph{Takeaway.}
On LoCoMo, PRISM is primarily anchor- and rerank-dominant: dense
anchor retrieval finds most relevant evidence, and Evidence
Compression compresses the candidate pool into a small focused
context without sacrificing accuracy. The graph components provide
the structural machinery needed for harder bridge-style settings,
but do not measurably affect this mostly single-evidence benchmark.
Adaptive Intent Routing further removes a substantial fraction of
classifier-side LLM calls without changing the accuracy profile.
 % =====================================================================
% SECTION 5 — CONCLUSION
% =====================================================================
\section{Conclusion}
\label{sec:conclusion}

We presented PRISM, a training-free retrieval-side framework for
long-horizon agent memory. Built on four orthogonal inference-time
modules---hierarchical bundle search over typed relation paths,
query-sensitive edge-cost adjustment, evidence compression, and
adaptive intent routing---PRISM formulates retrieval as a
min-cost selection over a graph-structured memory and pairs it with
an LLM-side compression step, enabling effective long-horizon
question answering at a small fraction of the context cost incurred
by full-history replay. Empirical results on the LoCoMo benchmark
demonstrate that PRISM achieves a strong balance between answer
accuracy and retrieval efficiency, occupying a previously empty
corner of the accuracy--context-cost frontier. These results
highlight the value of treating long-horizon memory as a joint
retrieval--compression problem, and suggest a promising direction
for building cost-efficient agents that scale to ever-longer
interaction histories. We discuss PRISM's limitations and broader impacts in Appendix~\ref{app:limitations-impacts}.

\bibliographystyle{plain}     
\bibliography{references}  
% ======================  附录 ======================

\clearpage
 \appendix
 % =====================================================================
% appendix.tex
%
% Usage in main:
%   \appendix
%   \input{appendix.tex}
%
% Sections:
%   A. PRISM Inference Algorithm
%   B. Implementation Details
%   C. Prompts
%   D. Statistical and Diagnostic Analyses
% =====================================================================

% Keep short appendix pages compact rather than vertically stretched by flushbottom.
\raggedbottom

% =====================================================================
% A. Limitations and Broader Impacts
% =====================================================================
\section{Limitations and Broader Impacts}
\label{app:limitations-impacts}

\begingroup
\setlength{\parskip}{0pt}
\noindent\textbf{Limitations.}
PRISM currently focuses on retrieval-side compression for LLM-based
long-horizon conversational memory. Future work can extend the same
principle to long agent trajectories involving actions, tool calls,
observations, plans, and feedback. Because PRISM is a training-free
retrieval-side plug-in over structured memory, it can naturally adapt to
richer trajectory-level memory graphs as agent memory pipelines evolve.

\smallskip
\noindent\textbf{Broader impacts.}
PRISM may have positive societal impact by reducing the API and serving
cost of memory-augmented agents, making long-horizon memory systems more
accessible to research groups and applications that cannot afford
full-history replay. The main potential negative impact is that PRISM
faithfully retrieves evidence stored by the upstream ingestion pipeline;
biases, sensitive content, or privacy-relevant information in the
underlying memory graph may therefore be reflected in the retrieved
context. Deployments should pair PRISM with ingestion-side filtering,
access controls, and application-specific privacy and fairness checks.
\endgroup

% =====================================================================
% A. PRISM Inference Algorithm
% =====================================================================
\section{PRISM Inference Algorithm}
\label{app:algorithm}

This section provides the full pseudocode for the PRISM inference
pipeline referenced in Section~\ref{sec:pipeline}.
Algorithm~\ref{alg:prism} makes explicit the four inference-time
modules---N4 adaptive intent routing, N1 typed-path candidate
generation under N2 intent-conditioned edge costs, and N3 LLM-side
evidence compression---and the equations governing each step.

% -------------------------------------------------------------------------
% Algorithm 1 -- PRISM end-to-end inference pipeline
% Requires: \usepackage{algorithm} and \usepackage{algpseudocode} or
% \usepackage{algorithmicx} in main.tex. We use algorithm + algpseudocode.
% -------------------------------------------------------------------------
\begin{algorithm}[H]
\caption{PRISM Inference Pipeline}
\label{alg:prism}
\begin{algorithmic}[1]
\Require Query $q$ with embedding $\mathbf{q}$; memory graph
         $\mathcal{G} = (\mathcal{V}, \mathcal{E}, \tau)$; layer-specific
         FAISS indices $\{F_\ell\}$; path-template set $\mathcal{T}$
         (\S\ref{sec:n1}); bundle size $K{=}10$; rerank size $M{=}5$.
\Ensure  Compact answer-side context $C$ of $\le M$ Episode summaries.
\Statex
\State \textbf{// N4: adaptive intent routing}
\State $h(q) \gets \Call{KeywordGate}{q}$
       \Comment{regex triggers; zero LLM calls}
\If{$h(q) = \bot$}
    \State $h(q) \gets \Call{PrototypeMatch}{\mathbf{q},\, \theta_{\text{proto}}}$
           \Comment{embedding cosine; zero LLM calls}
\EndIf
\If{$h(q) = \bot$}
    \State $h(q) \gets \Call{LlmClassify}{q}$
           \Comment{single classification call}
\EndIf
\Statex
\State \textbf{// N1 Phase 1: anchor discovery}
\State $\mathcal{A} \gets \bigcup_{\ell \in \{\text{Ep},\text{Fc},\text{FP},\text{En}\}}
        \Call{FaissTopK}{F_\ell,\, \mathbf{q}}$
\For{each anchor $a \in \mathcal{A}$}
    \State $d(a) \gets 1 - \cos(\mathbf{q}, \mathbf{e}_a)$
           \Comment{anchor cost}
\EndFor
\Statex
\State \textbf{// N1 Phases 2--3: typed path enumeration and costing (N2 inside)}
\State $\mathcal{P} \gets \emptyset$
\For{each $a \in \mathcal{A}$ and each template $t \in \mathcal{T}$}
    \For{each instantiation $\pi = (a, e_1, v_1, \ldots, e_h, \mathrm{Ep})$ of $t$}
        \State $\mathrm{Cost}(\pi) \gets d(a) + \sum_{i=1}^{h}
                \bigl(c_0(\tau(e_i)) \cdot
                      \alpha(\tau(e_i),\, h(q)) + c_{\text{hop}}\bigr)$
               \Comment{Eqs.~\ref{eq:path-cost},\,\ref{eq:edge-cost}}
        \State $\mathcal{P} \gets \mathcal{P} \cup \{\pi\}$
    \EndFor
\EndFor
\Statex
\State \textbf{// N1 Phase 4: bundle assembly via per-Episode minimum}
\For{each Episode $\mathrm{Ep}$ reached by some $\pi \in \mathcal{P}$}
    \State $s(\mathrm{Ep}) \gets \min_{\pi \in \Pi(\mathrm{Ep})} \mathrm{Cost}(\pi)$
\EndFor
\State $\mathcal{B} \gets \Call{TopK}{\{\mathrm{Ep} : s(\mathrm{Ep})\}_{\text{ascending}},\, K}$
       \Comment{candidate bundle, $|\mathcal{B}| = 10$}
\Statex
\State \textbf{// N3: evidence compression}
\State $(\mathrm{Ep}_{(1)}, \ldots, \mathrm{Ep}_{(M)}) \gets
        \Call{LlmReRank}{q,\, \mathcal{B},\, M}$
       \Comment{single LLM call; content-only prompt}
\State $C \gets \Call{Concat}{\mathrm{Summary}(\mathrm{Ep}_{(1)}),
                              \ldots,
                              \mathrm{Summary}(\mathrm{Ep}_{(M)})}$
\State \Return $C$
\end{algorithmic}
\end{algorithm}

% =====================================================================
% B. Implementation Details
% =====================================================================
\section{Implementation Details}
\label{app:implementation}

This section provides additional implementation details that
complement the description in Section~\ref{sec:method}.

% --------------------------------------------------------------------
% --------------------------------------------------------------------
\subsection{Memory Graph Construction}
\label{app:graph-construction}

This section describes the offline ingestion pipeline that converts a
raw LoCoMo conversation into the multi-relational memory graph
$\mathcal{G} = (\mathcal{V}, \mathcal{E}, \tau)$ defined in
Section~\ref{sec:graph}. Ingestion is performed once per conversation
and produces a reusable checkpoint. All same-protocol experiments in
Table~\ref{tab:locomo_main} reuse the same ingested graph, so that
ablations differ only in retrieval-side configuration.

\paragraph{Pipeline overview.}
A conversation is first split into chunks. Each chunk is then processed
by a six-stage pipeline:
SHA-256 deduplication
$\to$ LLM-based structured extraction
$\to$ graph node and edge construction
$\to$ embedding and FAISS indexing
$\to$ asynchronous causal consolidation
$\to$ checkpoint write. The pipeline is monotonic: once a node or edge
is written to $\mathcal{G}$, it is not deleted during ingestion.
Identical chunks are skipped through hash-based deduplication.

\paragraph{Structured extraction.}
For each chunk, we issue a single LLM call using
\texttt{gpt-4o-mini} with temperature $0.0$. The model returns a
strongly typed JSON object with five fields. The first field is an
Episode summary that covers the events and facts in the chunk. The
second is a list of Entities with type labels, such as person,
organization, place, concept, event, or other. The third is a list of
FacetPoints, where each FacetPoint represents one atomic fact and may
carry a related Entity name and a temporal expression. The fourth is a
list of Facets, where each Facet groups thematically related
FacetPoints by index. The fifth is a list of temporal annotations,
normalised to ISO-8601 when explicit; otherwise, the timestamp is
resolved by a best-effort absolute-date estimate from the chunk-header
conversation time. The extraction prompt is reproduced in
Appendix~\ref{app:prompt-extraction}. The output JSON is validated
against a Pydantic schema. If validation fails or the output is empty,
a fallback extractor produces only the Episode summary and Entity list,
which prevents a single malformed chunk from aborting ingestion.

\paragraph{Node construction and merging.}
Nodes are inserted layer by layer. Each Episode is added as one node
and inserted into a chronological chain according to the chunk-header
timestamp. Each Entity is deduplicated against existing Entity nodes
using exact name matching followed by cosine similarity over
entity-name embeddings; matches above $0.90$ are merged into the
existing node. Each FacetPoint is added as a fresh node linked back to
its Episode and, when a related Entity is available, connected to that
Entity through an \texttt{involves\_entity} shortcut edge. Each Facet
is merged against existing Facets within the same Episode when their
embedding cosine similarity exceeds $0.85$; otherwise, a new Facet node
is created and linked to its constituent FacetPoints. The four-layer
hierarchy
\texttt{Entity}--\texttt{FacetPoint}--\texttt{Facet}--\texttt{Episode}
is therefore populated within each chunk, with greedy cross-chunk
deduplication applied to the Entity layer and within-Episode merging
applied to the Facet layer.

\paragraph{Hierarchical and relation edges.}
Hierarchical \texttt{belongs\_to} edges are written eagerly during node
construction:
Entity\,$\to$\,FacetPoint\,$\to$\,Facet\,$\to$\,Episode. Of the five
relation edge types defined in Section~\ref{sec:graph}, three are
written synchronously during the same-chunk build phase, one is
deferred to an asynchronous consolidation pass, and one is disabled in
the LoCoMo configuration used in our experiments. Specifically,
\textsc{temporal} edges connect FacetPoints whose extracted temporal
annotations admit a definite \emph{before}/\emph{after} ordering.
\texttt{involves\_entity} edges are the FacetPoint--Entity shortcut
edges described above. \textsc{evolution} edges connect FacetPoints
across chunks that share a related Entity by linking the temporally
latest existing FacetPoint of that Entity to the temporally earliest
new FacetPoint introduced by the current chunk. Within a chunk,
multiple FacetPoints sharing the same Entity are linked pairwise in
temporal order. \textsc{causal} edges are deferred to the asynchronous
consolidation pass described next. \textsc{semantic} edges are defined
in our schema as cosine-similarity-based cross-document links between
conceptually related Facets, but are disabled by default in the LoCoMo
configuration used for our experiments. In early pilot runs, we
observed that on conversational data with LoCoMo's density,
similarity-induced semantic edges produced many weak links that
increased candidate noise without measurable retrieval gains. We
therefore retain the edge type in the graph schema but do not
instantiate it during ingestion. We expect \textsc{semantic} edges to
be more useful on denser corpora with richer cross-document conceptual
overlap, and leave their re-activation to future work.

\paragraph{Asynchronous causal consolidation.}
Every $5$ ingested chunks, the pipeline launches an asynchronous
causal-consolidation pass. The pass collects the $5$ most recent
Episodes and issues a second LLM call using \texttt{gpt-4o-mini} with
temperature $0.0$. The model returns a list of
$\langle\text{cause}, \text{effect}, \text{description},
\text{confidence}\rangle$ tuples. Pairs with confidence below $0.7$
are discarded; pairs whose endpoint IDs are missing from the graph or
whose corresponding \textsc{causal} edge already exists are also
dropped. Surviving pairs are written as \textsc{causal} edges, with
the LLM-assigned confidence retained as edge metadata. This deferred
design lets the model observe a small window of recent Episodes at
once, which improves cross-Episode causal precision compared with a
single-chunk extraction call, while the confidence floor controls the
false-positive rate of LLM-induced causal links.

\paragraph{Embedding and FAISS indexing.}
All non-\texttt{belongs\_to} edges are embedded together with their
text descriptions and registered in one of two FAISS edge indices.
\textsc{semantic} edges, when enabled, are written to
\texttt{edge\_semantic}; \textsc{temporal}, \textsc{causal},
\textsc{evolution}, and \texttt{involves\_entity} edges are written to
\texttt{edge\_relation}. In the LoCoMo configuration used for our
experiments, \texttt{edge\_semantic} is therefore empty, while
\texttt{edge\_relation} stores the four active relation edge types.
Nodes at each of the four hierarchy layers are indexed in their own
per-layer FAISS index. All embeddings use
\texttt{all-MiniLM-L6-v2} with dimension $d=384$ and $L_2$
normalisation, after which \texttt{IndexFlatIP} inner-product search
is equivalent to cosine retrieval. The ingestion and indexing
hyperparameters, including merge thresholds and consolidation
interval, are summarised in Appendix~\ref{app:hyperparams}.

\paragraph{Checkpoint and reuse.}
The ingestion pipeline persists three artefacts: the typed graph
$\mathcal{G}$ stored as \texttt{graph.json}, the FAISS indices stored
under \texttt{faiss/}, and the raw chunk text store stored as
\texttt{chunks.json}. A small bookkeeping file records the set of
ingested chunk hashes for deduplication. All experiments in
Section~\ref{sec:expert} reuse a single ingest checkpoint per LoCoMo
conversation, ensuring that differences across ablations come from
retrieval-side components rather than from graph construction.

% --------------------------------------------------------------------
\subsection{Path Templates for Hierarchical Bundle Search}
\label{app:path-templates}

Hierarchical Bundle Search enumerates eight typed path templates,
partitioned into a backbone family and a relation-bridge family.
Backbone templates traverse only \texttt{belongs\_to} edges; the
three relation-bridge templates each cross exactly one typed relation
edge before re-entering the hierarchy.

\begin{quote}
\small
\begin{verbatim}
Backbone (5 templates):
    Ep
    Fc -> Ep
    FP -> Fc -> Ep
    En -> FP -> Fc -> Ep
    En -> Fc -> Ep

Relation-bridge (3 templates):
    a --temporal-->  v -> Ep
    a --causal---->  v -> Ep
    a --evolution--> v -> Ep
\end{verbatim}
\end{quote}

\noindent
Here \texttt{Ep}, \texttt{Fc}, \texttt{FP}, and \texttt{En} denote
Episode, Facet, FacetPoint, and Entity nodes respectively, and
\texttt{a} is the query anchor.  Backbone paths capture evidence
whose relevance is established by similarity to the anchor;
relation-bridge paths surface indirect evidence whose target chunk
shares no surface similarity with the query, making them the
structural basis for multi-hop and causal question answering.

% --------------------------------------------------------------------
\subsection{Hyperparameter Configuration}
\label{app:hyperparams}

Table~\ref{tab:hyperparams} reports the hyperparameter values used
for all PRISM experiments. Values were fixed before any test-set
evaluation and held constant across all rows of
Table~\ref{tab:locomo_main} and Table~\ref{tab:ablation}.

\begin{table}[h]
    \centering
    \small
    \caption{Hyperparameter configuration for PRISM. All values are
    fixed across LoCoMo categories~1--4 and held identical across
    the same-protocol rows of Table~\ref{tab:locomo_main}.}
    \label{tab:hyperparams}
    \begin{tabular}{@{}llc@{}}
        \toprule
        Module & Parameter & Value \\
        \midrule
        \multicolumn{3}{l}{\emph{Ingestion and Indexing}} \\
        & Embedding model                          & \texttt{all-MiniLM-L6-v2} \\
        & Embedding dimension $d$                  & $384$ \\
        & Vector index                             & FAISS (\texttt{IndexFlatIP}, $L_2$-normalised) \\
        & Entity merge threshold                   & $0.90$ \\
        & Facet merge threshold                    & $0.85$ \\
        & Causal consolidation interval            & $5$ chunks \\
        \midrule
        \multicolumn{3}{l}{\emph{N1: Hierarchical Bundle Search}} \\
        & FAISS top-$k$ (all four layers)          & $30$ \\
        & Hop penalty $c_{\text{hop}}$             & $0.05$ \\
        & Bundle size $K$                          & $10$ \\
        \midrule
        \multicolumn{3}{l}{\emph{N2: Query-Sensitive Edge Cost (Eq.~\ref{eq:edge-cost})}} \\
        & In-recall edge cost                      & $1 - \cos(\mathbf{e}_{e}, \mathbf{q})$ \\
        & Edge-miss fallback $c_0(\textsc{belongs\_to})$    & $0.02$ \\
        & Edge-miss fallback $c_0(\textsc{semantic})$       & $0.90$ \\
        & Edge-miss fallback $c_0(\textsc{temporal})$       & $0.90$ \\
        & Edge-miss fallback $c_0(\textsc{causal})$         & $0.90$ \\
        & Edge-miss fallback $c_0(\textsc{evolution})$      & $0.90$ \\
        & Discount $\alpha$ (matched, $\tau \in \{\textsc{temporal}, \textsc{causal}\}$) & $0.5$ \\
        & Discount $\alpha$ (\textsc{evolution} under \textsc{temporal} intent)          & $0.7$ \\
        & Discount $\alpha$ (otherwise)                                                   & $1.0$ \\
        \midrule
        \multicolumn{3}{l}{\emph{N3: Evidence Compression}} \\
        & Rerank model                             & \texttt{gpt-4o-mini} \\
        & Rerank temperature                       & $0.0$ \\
        & Rerank size $M$                          & $5$ \\
        & Snippet truncation                       & $400$ chars \\
        \midrule
        \multicolumn{3}{l}{\emph{N4: Adaptive Intent Routing}} \\
        & Keyword bank size$^{\dagger}$            & $32$ \\
        & Prototype bank size$^{\ddagger}$         & $46$ \\
        & Prototype confidence threshold $\theta_{\text{proto}}$ & $0.55$ \\
        & Prototype top-$2$ margin                 & $0.10$ \\
        & LLM fallback model                       & \texttt{gpt-4o-mini} \\
        \midrule
        \multicolumn{3}{l}{\emph{Answer / Judge}} \\
        & Answer model (default)                   & \texttt{gpt-4o-mini} \\
        & Answer temperature                       & $0.0$ \\
        & Judge model                              & \texttt{gpt-4o-mini} \\
        & Judge temperature                        & $0.0$ \\
        & Random seed                              & $42$ \\
        \bottomrule
    \end{tabular}
\end{table}

\noindent
\footnotesize
$^{\dagger}$ Combines $18$ \textsc{temporal} triggers (\eg, \emph{when}, \emph{before}, \emph{how long}) and $14$ \textsc{causal} triggers (\eg, \emph{why}, \emph{because}).
$^{\ddagger}$ Combines $24$ generic prototypes and $22$ LoCoMo-specific prototypes, distributed across intent labels as \textsc{temporal}: $10$, \textsc{causal}: $10$, \textsc{multi\_hop}: $14$, \textsc{entity\_centric}: $12$. The prototype path commits a label only when the top-$1$ prototype similarity exceeds $\theta_{\text{proto}}$ and the top-$2$ margin exceeds $0.10$; otherwise the query falls through to the LLM classifier.
\normalsize

% =====================================================================
% C. Prompts
% =====================================================================
\section{Prompts}
\label{app:prompts}

For full reproducibility we list the prompts used in PRISM. All
prompts are model-agnostic; in our experiments the answer model,
the N3 reranker, the N4 LLM-fallback classifier, and the LLM judge
all use \texttt{gpt-4o-mini} at \texttt{temperature}=$0.0$. The
answer and judge prompts follow the answer-then-judge protocol
used in the public Mem0 memory-benchmarks LoCoMo
evaluation~\citep{chhikara2025mem0}, adapted with project-specific
instructions. All prompts are reproduced verbatim below; \texttt{\{...\}}
denotes runtime template fields.

% --------------------------------------------------------------------
\subsection{Schema-Guided Extraction Prompts}
\label{app:prompt-extraction}

The ingestion pipeline (Appendix~\ref{app:graph-construction})
applies two LLM-side prompts: one to extract the four-layer
hierarchy (Episode, Facet, FacetPoint, Entity) together with
temporal annotations from each conversation chunk, and one to infer
causal edges from a list of episode summaries.

\paragraph{Hierarchy and Temporal Extraction.}
This prompt produces the Episode summary, Entity nodes, FacetPoint
nodes, Facet groupings, and temporal annotations attached to each
chunk. Output is constrained to a strict JSON schema; no markdown
or commentary is permitted.

\begin{quote}
\scriptsize
\begin{verbatim}
You are an information extraction engine.

Task:
Extract structured memory data from the text chunk below.

Text chunk:
{chunk}

Return a single JSON object with exactly these top-level keys:
- episode_summary: string
- entities: array of objects
- facet_points: array of objects
- facets: array of objects
- temporal_info: array of objects

Schema details:
1) episode_summary
- A concise but comprehensive summary of ALL events and facts mentioned
  in the chunk. Include specific details like names, dates, places,
  objects, and quantities.

2) entities
- Each item must be:
    {"name": string, "entity_type": string}
- entity_type should be one of: "person", "organization", "place",
  "concept", "event", "other".
- Keep names as they appear in the text whenever possible.
- Include specific items mentioned (books, foods, activities, pets,
  places visited, etc.) as entities with type "concept" or "other".

3) facet_points
- Each item must be:
    {"content": string,
     "related_entity_name": string or null,
     "timestamp_text": string or null}
- content should be atomic and factual.
- IMPORTANT: Be specific. Include concrete details like exact names,
  quantities, colors, and descriptions.
    Good: "Melanie made a cup in her pottery class"
    Bad : "Melanie does pottery"
    Good: "Caroline recommended the book Becoming Nicole to Melanie"
    Bad : "Caroline recommended a book"
- Extract every distinct fact as a separate facet_point. Do not merge
  multiple facts into one.

4) facets
- Each item must be:
    {"theme": string, "facet_point_indices": array of integers}
- facet_point_indices refers to zero-based indices in the
  facet_points array.

5) temporal_info
- Each item must be:
    {"subject": string,
     "time_expression": string,
     "normalized_time": string or null,
     "relation": string}
- relation examples: "before", "after", "during", "at".
- normalized_time should use ISO-8601 when explicit enough,
  otherwise null.
- For relative time references (e.g., "yesterday", "last week"),
  use the conversation timestamp shown in the chunk header to
  compute an absolute date for normalized_time.

Rules:
- Output valid JSON only. No markdown fences, no explanation text.
- If information is missing, use empty arrays or null values
  as appropriate.
- Do not invent unsupported facts.
- Prefer extracting MORE facet_points with specific details over
  fewer generic ones.
\end{verbatim}
\end{quote}

\paragraph{Causal Edge Induction.}
After Episode-level extraction, a second prompt operates on a list
of episode summaries to populate the \textsc{causal} relation edges
of $\mathcal{G}$. Each induced edge carries an LLM-assigned
confidence score in $[0, 1]$, retained as edge metadata.

\begin{quote}
\scriptsize
\begin{verbatim}
You are a causal reasoning engine.

Task:
Given a list of episode summaries, identify likely causal relations.

Events:
{events}

Return a single JSON object with exactly one top-level key:
- causal_pairs: array of objects

Each causal_pairs item must be:
{"cause_id": string,
 "effect_id": string,
 "description": string,
 "confidence": number}

Rules:
- cause_id and effect_id must refer to event IDs provided in the input.
- confidence must be between 0.0 and 1.0.
- Keep descriptions concise and evidence-aware.
- Only include relations with meaningful support from the provided
  events.
- Output valid JSON only. No markdown fences, no explanation text.
- If no reliable causal relation exists, return:
    {"causal_pairs": []}.
\end{verbatim}
\end{quote}

% --------------------------------------------------------------------
% --------------------------------------------------------------------
\subsection{N3 Reranker Prompt}
\label{app:prompt-rerank}

The N3 module (Eq.~\ref{eq:compress}) is implemented as a
score-then-select pipeline: the LLM is asked to assign a relevance
score in $[0, 10]$ to each candidate Episode summary, after which
the top-$M$ scoring Episodes are selected deterministically. The
prompt is content-only---it sees Episode summaries but receives
neither path-cost scores nor bundle metadata---so its judgement is
independent of the structural signal that produced the candidate
bundle. Each Episode summary is truncated to the first $400$
characters before being shown to the reranker, keeping the prompt
length bounded for typical bundle sizes ($K = 10$). Output is
constrained to a JSON array of \texttt{(index, score)} pairs; the
deterministic selection of the top-$M$ Episodes is performed by
post-processing in our retrieval layer.

\begin{quote}
\scriptsize
\begin{verbatim}
Rate the relevance of each memory snippet to the question on a
scale of 0-10.

Question: {question}

Snippets:
{snippets}

Return a JSON array of objects, each with "index" (integer) and
"score" (0-10).
Example: [{"index": 0, "score": 9}, {"index": 1, "score": 2}]

Return ONLY the JSON array, no other text.
\end{verbatim}
\end{quote}

When the bundle size satisfies $|\mathcal{B}| \le M$, this stage is
skipped and all candidates are forwarded to the answer model
unchanged; reranking is only invoked when the candidate bundle
exceeds the answer-side budget.

% --------------------------------------------------------------------
\subsection{Answer and LLM-Judge Prompts}
\label{app:prompt-judge}

The answer model receives the top-$M$ Episode summaries selected by
N3 (Eq.~\ref{eq:compress}) and produces a free-form answer; the
LLM judge then compares this answer to the gold reference and emits
a binary CORRECT/WRONG label. Both prompts are project-specific and
follow the answer-then-judge style of the public Mem0
memory-benchmarks LoCoMo evaluation.

\paragraph{Answer Prompt.}
\begin{quote}
\scriptsize
\begin{verbatim}
You are a helpful assistant answering questions based on the
provided context.

Context:
{context}

Question: {question}

Instructions:
1. Answer the question using the provided context. Be specific and
   cite concrete details from the context.
2. For time-related questions, follow these steps:
   Step 1: Find the conversation date from the header
           (e.g., [1:56 pm on 8 May, 2023] means the conversation
           date is 8 May 2023).
   Step 2: Identify the relative time expression
           (e.g., "yesterday", "last week", "last Saturday").
   Step 3: Calculate the actual date.
           "yesterday" = conversation date minus 1 day.
           "last week" = approximately 7 days before.
           "last Saturday" = the most recent Saturday before the
                             conversation date.
           "next month" = the month after the conversation month.
   Step 4: State the calculated date in your answer.
3. When multiple events of the same type exist (e.g., multiple
   camping trips, multiple beach visits), distinguish between them
   using their dates.
4. Prefer quoting specific details (names, dates, objects, places)
   from the context over paraphrasing.
5. If the context contains partial but relevant information,
   provide the best answer you can.
6. Only say you cannot answer if the context truly contains NO
   relevant information at all.

Answer:
\end{verbatim}
\end{quote}

\paragraph{LLM-as-a-Judge Prompt.}
\begin{quote}
\scriptsize
\begin{verbatim}
You are an evaluation judge. Compare the generated answer with the
gold answer and determine if the generated answer is correct.

Be lenient with format differences. For example:
- "May 7th" and "7 May" are the same date  ->  CORRECT
- "Caesar salad" and "She ordered a Caesar salad"  ->  CORRECT
                                                       (same meaning)
- Partial but accurate answers are CORRECT

Question: {question}
Gold answer: {gold_answer}
Generated answer: {generated_answer}

First, provide a short (one sentence) explanation of your reasoning,
then finish with CORRECT or WRONG.

Return your response in JSON format:
{"reasoning": "your explanation", "label": "CORRECT or WRONG"}
\end{verbatim}
\end{quote}

% --------------------------------------------------------------------
\subsection{Adaptive Intent Routing LLM Prompt}
\label{app:prompt-intent}

This is the prompt used by the third-tier LLM fallback in the
Adaptive Intent Routing cascade (Eq.~\ref{eq:cascade}), invoked
only when both the keyword-gated and prototype-matching tiers
return no confident label. The prompt asks the LLM to score each
of the four intent types independently in $[0, 1]$; the downstream
routing layer then thresholds these scores and applies the
\textsc{multi\_hop}-suppresses-\textsc{entity\_centric} rule
described in \S\ref{sec:n4}, with \textsc{general} assigned when no
score exceeds the routing threshold.

\paragraph{System Prompt.}
\begin{quote}
\scriptsize
\begin{verbatim}
You are an intent classifier for a memory retrieval system. Your
job is to analyze a user's query and estimate how strongly it
expresses each of four intent types. Output only valid JSON. Do
not add commentary.
\end{verbatim}
\end{quote}

\paragraph{User Prompt Template.}
\begin{quote}
\scriptsize
\begin{verbatim}
Analyze the intent of the following query. A query may express
multiple intents simultaneously.

Query: {query}

Intent types and what counts as a signal for each:

1. temporal -- The query asks about WHEN something happened, time
   ordering, duration, or sequence of events. Signals: "when",
   "before", "after", "during", "how long", "what year", explicit
   dates, or asking about the timing of events relative to each
   other.

2. causal -- The query asks WHY something happened, what caused
   it, or what led to an outcome. Signals: "why", "because",
   "what caused", "what led to", "as a result of", or asking
   about reasons, motivations, or consequences.

3. multi_hop -- The query requires combining facts from multiple
   separate events, interactions, or contexts to answer. A
   single-fact lookup is NOT multi_hop. Signals: "based on X
   and Y", "how does X relate to Y", "given that ... what ...",
   "combining these conversations", "across multiple sessions",
   asking about trends/patterns/shifts across time, or asking for
   inferences that require connecting separate pieces of
   information. Also includes hypothetical, predictive, or
   inference-requiring questions where the answer must be derived
   rather than directly retrieved -- even when the subject is a
   named person, the inference requirement makes this multi_hop,
   NOT entity_centric.

4. entity_centric -- The query asks about a specific attribute,
   description, or property of a person, place, or thing that can
   be looked up as a stored fact. Signals: "who is", "what does
   X look like", "where does X live", "what is X's job", or
   asking to retrieve a single concrete fact about a named
   entity. NOTE: if answering requires inference or reasoning
   across multiple facts rather than a direct lookup, score
   multi_hop higher than entity_centric.

Rules:
- Score each intent independently on a scale of 0.0 to 1.0.
- Multiple intents can score high at once (e.g., a query can be
  both temporal and multi_hop).
- If the query clearly expresses no recognizable intent, all four
  scores should be low (< 0.3).
- For vague conversational continuations ("Tell me more",
  "Continue", "Go on"), all scores should be 0.0 -- do not guess
  at entity or any other intent.
- Prefer 0.0 / 0.3 / 0.6 / 0.9 as anchor points when in doubt.

Output format (JSON object, no other text):
{"temporal": 0.X, "causal": 0.X, "multi_hop": 0.X, "entity_centric": 0.X}
\end{verbatim}
\end{quote}

% =====================================================================
% D. Statistical and Diagnostic Analyses
% =====================================================================
\section{Statistical and Diagnostic Analyses}
\label{app:diagnostics}

This section provides the statistical and diagnostic analyses
referenced in Section~\ref{sec:ablation}: full paired-bootstrap
confidence intervals for every ablation row
(\S\ref{app:ci_full}); the manual annotation study used to
diagnose the null result of the relation paths (N1) and edge
costs (N2) on LoCoMo (\S\ref{app:multihop_anno}); and the
per-category routing distribution of the optional adaptive intent
routing module (N4) (\S\ref{app:routing_per_category}).

% --------------------------------------------------------------------
% --------------------------------------------------------------------
\subsection{Full Paired-Bootstrap Confidence Intervals}
\label{app:ci_full}

We report paired-bootstrap $95\%$ confidence intervals and
McNemar mid-$p$ values for every ablation row in
Table~\ref{tab:ablation}, computed against PRISM on the same
$1{,}540$ LoCoMo cat~1--4 QA. CIs are obtained with $2{,}000$
paired resamples (seed $=42$); McNemar is two-sided mid-$p$. Each
row reports the discordant counts $(b, c)$, where $b$ is the
number of QA on which PRISM is correct and the variant is wrong,
and $c$ is the converse. As shown in
Table~\ref{tab:ablation_ci}, all four ablations produce
judge-score CIs that include zero, indicating that none reaches
statistical significance at $\alpha = 0.05$ on the judge metric;
$-$N3 shows the largest absolute judge-score gap
($-0.58$\,pp) but remains well within bootstrap noise. The
component-level effects of $-$N3 reported in
Table~\ref{tab:ablation}---a $-6.8$\,pp drop in ER@$5$ and a
$+2{,}085$-token increase in answer-side context---therefore
correspond to changes in \emph{which} evidence reaches the top of
the context and \emph{how much} context is forwarded to the
answer model, rather than to a measurable shift in the binary
judge label.

\begin{table}[h]
    \centering
    \small
    \setlength{\tabcolsep}{4pt}
    \caption{Full paired-bootstrap $95\%$ CIs and McNemar mid-$p$
    values for the four ablations of Table~\ref{tab:ablation},
    computed against PRISM on $1{,}540$ LoCoMo cat~1--4 QA.
    $\Delta$ is in percentage points (variant minus PRISM); $b$
    and $c$ are the discordant counts. None of the four CIs
    excludes zero on judge accuracy.}
    \label{tab:ablation_ci}
    \begin{tabular}{@{}lrrrcc@{}}
        \toprule
        Variant & $b$ & $c$ & $\Delta$ Judge (pp)
                & McNemar mid-$p$ & $95\%$ CI (pp) \\
        \midrule
        $-$N1 (no relation paths)   & $43$ & $43$ & $\phantom{+}0.00$ & $0.957$ & $[-1.17,\ +1.23]$ \\
        $-$N2 (no cost adjustment)  & $49$ & $50$ & $+0.06$            & $0.960$ & $[-1.23,\ +1.30]$ \\
        $-$N3 (no LLM rerank)       & $76$ & $67$ & $-0.58$            & $0.479$ & $[-2.08,\ +0.91]$ \\
        $+$N4 (hybrid intent)       & $42$ & $46$ & $+0.26$            & $0.711$ & $[-0.91,\ +1.43]$ \\
        \bottomrule
    \end{tabular}
\end{table}

% --------------------------------------------------------------------
\subsection{Multi-Hop Annotation Study}
\label{app:multihop_anno}

To diagnose why the relation-paths and edge-cost components
(N1, N2) produce a null result on LoCoMo
(\S\ref{sec:ablation}), we manually annotated a random sample of
$50$ multi-hop questions from the $282$ multi-hop QAs in
categories~1--4.

\paragraph{Sampling.}
We restrict the population to multi-hop questions because
single-hop and open-domain categories are structurally
inapplicable to N1 and N2 by design. Of the $282$ multi-hop QAs
in cat~1--4, we drew a uniform random sample of $50$ without
replacement, with seed $=42$ fixed before any analysis.

\paragraph{Labelling Criteria.}
For each sampled question $(q, A_q, E_q)$, where $E_q$ is the
gold \texttt{evidence} list, two annotators independently
labelled it along two axes:
\begin{itemize}
    \item \textbf{Anchor-discoverable}: every gold evidence turn
          $e \in E_q$ shares either a named entity or a topical
          keyword with the question $q$.
    \item \textbf{Bridge-style}: at least two gold turns
          $e_1, e_2 \in E_q$ are connected only by a shared
          \emph{intermediate} entity $z$ that does not appear in
          $q$---the canonical two-hop bridge that N1's
          relation-bridge paths and N2's causal/temporal
          discounts target.
\end{itemize}
Disagreements were resolved by discussion. Following the LoCoMo
convention, each list entry of \texttt{evidence} counts as one
citation; semicolon-joined compound dia-IDs ($\approx 2\%$ of
entries) are treated as a single citation per dataset
convention.

\paragraph{Results.}
$36$ of $50$ ($72\%$) questions are fully
anchor-discoverable. Only $3$ of $50$ ($6\%$, Wilson $95\%$ CI
$[2.1\%,\,16.5\%]$) require the two-hop bridge structure that N1
and N2 are designed to support. Extrapolating to the full
$282$-question multi-hop subset, the bridging cases correspond to
approximately $17$ of $282$ multi-hop QAs ($95\%$ CI $[6, 47]$),
or $\lesssim 3\%$ of the full $1{,}540$ benchmark.

\paragraph{Population Context.}
A separate count over the full $1{,}540$ cat~1--4 questions shows
that $73.4\%$ cite a single \texttt{evidence} entry and therefore
admit no cross-turn bridge by construction; the $50$-QA bridging
analysis above operates within the remaining $26.6\%$
multi-evidence subset. Together, these counts indicate that
LoCoMo cat~1--4 is a mostly anchor-discoverable benchmark and
that the null result for N1 and N2 reflects benchmark structure
rather than a defect of the modules themselves; we expect both
modules to contribute on benchmarks with lexically mismatched
multi-hop evidence such as MuSiQue~\citep{trivedi2022musique} or
bridge-style HotpotQA~\citep{yang2018hotpotqa}.

% --------------------------------------------------------------------
\subsection{Per-Category Routing Distribution of N4}
\label{app:routing_per_category}

\begin{figure}[h]
    \centering
    \includegraphics[width=0.85\linewidth]{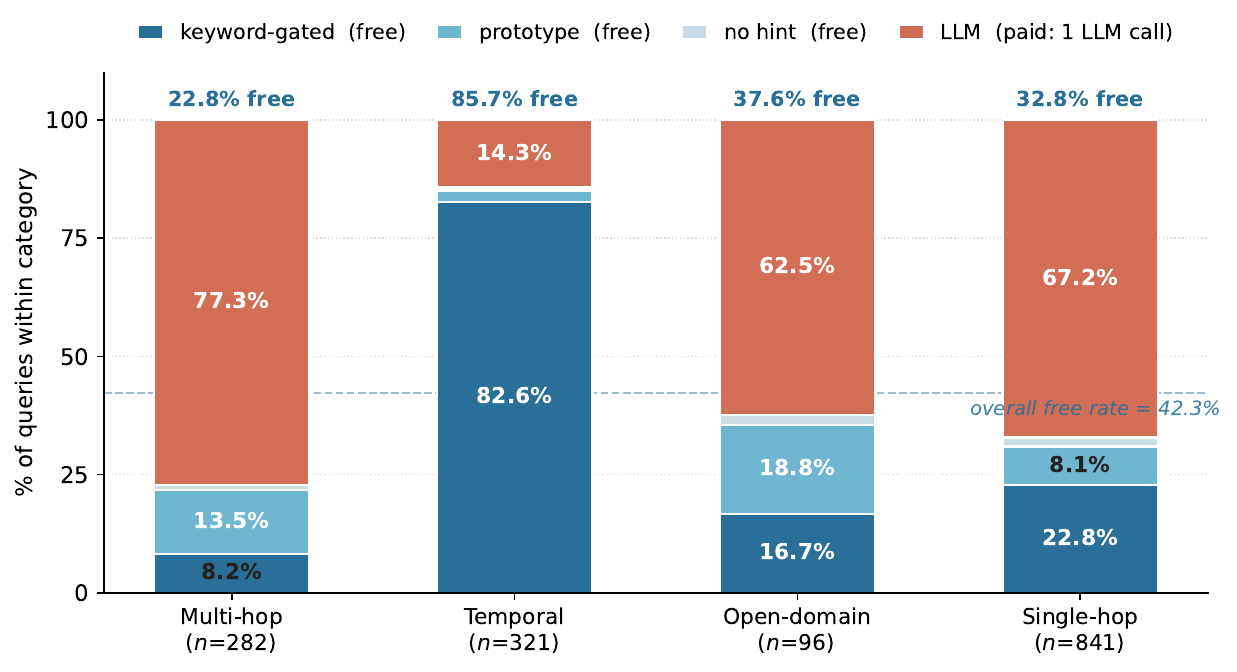}
    \caption{Per-category routing distribution of Adaptive Intent
    Routing (N4) on LoCoMo cat~1--4. Each bar shows the share of
    queries dispatched through each routing path. The
    \texttt{keyword\_gated}, \texttt{prototype}, and \texttt{none}
    paths incur zero LLM calls; only the \texttt{LLM} path incurs
    one classifier-side LLM call per query. The annotation marks
    the overall no-LLM rate of $42.3\%$.}
    \label{fig:routing_breakdown}
\end{figure}

Figure~\ref{fig:routing_breakdown} shows the per-category routing
distribution of Adaptive Intent Routing (N4) across the four
LoCoMo question types, complementing the aggregate breakdown in
Table~\ref{tab:routing}. The savings discussed in
\S\ref{sec:ablation} are highly non-uniform: temporal queries
route through keyword gating in $82.6\%$ of cases (the keyword
bank includes \emph{when}, \emph{before}, \emph{how long}, and
related triggers), while only $8$--$23\%$ of multi-hop,
single-hop, and open-domain queries match a keyword trigger; the
latter three categories therefore fall back to the LLM
classifier in $62$--$77\%$ of queries. The dashed annotation
marks the overall no-LLM rate of $42.3\%$ across the
$1{,}540$ QA. This per-category profile is consistent with the
no-accuracy-cost claim of Table~\ref{tab:paired}: the categories
where N4 saves most LLM calls (temporal) are also the categories
whose intent labels are most easily recovered without LLM
inference.

\end{document}